## Original Paper

Ravi P Garg, MSc[1], Kalpana Raja, PhD[1], Siddhartha R Jonnalagadda, PhD[1]*

[1]Division of Health and Biomedical Informatics, Department of Preventive Medicine,

Northwestern University Feinberg School of Medicine, Chicago, IL

# CRTS: A type system for representing clinical recommendations

## Abstract

**Background:** Clinical guidelines and recommendations are the driving wheels of the evidence-based medicine (EBM) paradigm, but these are available primarily as unstructured text and are generally highly heterogeneous in nature. This significantly reduces the dissemination and automatic application of these recommendations at the point of care. A comprehensive structured representation of these recommendations is highly beneficial in this regard.

**Objective:** The objective of this paper to present Clinical Recommendation Type System (CRTS), a common type system that can effectively represent a clinical recommendation in a structured form.

**Methods:** CRTS is built by analyzing 125 recommendations and 195 research articles corresponding to 6 different diseases available from UpToDate, a publicly available clinical knowledge system, and from the National Guideline Clearinghouse, a public resource for evidence-based clinical practice guidelines.

**Results:** We show that CRTS not only covers the recommendations but also is flexible to be extended to represent information from primary literature. We also describe how our developed type system can be applied for clinical decision support, medical knowledge summarization, and citation retrieval.

**Conclusion:** We showed that our proposed type system is precise and comprehensive in representing a large sample of recommendations available for various disorders. CRTS can now be used to build interoperable information extraction systems that automatically extract clinical recommendations and related data elements from clinical evidence resources, guidelines, systematic reviews and primary publications.

## Introduction

During the past decade, there has been an overwhelming increase in the amount of data generated in biomedical field. More than 700,000 biomedical primary studies are added to MEDLINE each year from 2009 [1]. It has become impossible for clinicians to keep track of each study published in their field of expertise. It is also known that medical errors that could have been prevented are leading cause of deaths in US patients [2] and this situation is accompanied by a continual rise in healthcare costs and increased complexity of diseases [3]. The recent advances and successes of information extraction systems such as Watson [4] and large amounts of publicly available data, information and knowledge provide a huge opportunity to mitigate these errors and control healthcare costs through new computational approaches to integrate clinical decision support (CDS) [5] and evidence-based medicine (EBM) [6]. CDS systems aim to assist physicians and healthcare professionals with the clinical decision-making process [7-10]. EBM refers to practice of medicine based on best available evidence from the literature [11, 12]. The integration of CDS with EBM thereby holds great promise and potential to handle the aforementioned healthcare problems.

## Clinical Knowledge Systems and Clinical Practice Guidelines

Clinical knowledge systems and clinical practice guidelines form the driving wheels of EBM. However, due to the recent explosion in the creation and availability of medical literature, maintaining and updating these systems take considerable time and manual effort.

UpToDate is an example of an evidence-based clinical knowledge system that is widely used by clinicians and healthcare professionals [13]. It provides extensive information related to several medical topics (diagnosis, therapy, prevention, etiology, management, etc.) for many disorders. For example, "Overview of the therapy of heart failure due to systolic dysfunction" is a topic for the disease heart failure (HF). Content for these topics are compiled by experts in those topics. Each topic describes a particular subject in detail before giving relevant conclusions in the form of recommendations. Recommendations are structured summaries of the evidence found in the relevant citations for each topic. An example of a recommendation is, "*For patients with systolic HF who do not tolerate ACE [angiotensin-converting enzyme] inhibitors, we recommend an angiotensin II receptor blocker (ARB) as an alternative that provides a similar survival benefit.*" We list some more examples of these recommendations in Table 1 and Table 2. The citations from which evidence was gathered are then provided as links for each recommendation.

These recommendations provide crucial information for clinicians and medical practitioners, helping them to find and take adequate action at the point of care. However, because the recommendations are unstructured and use heterogeneous style of language, representing such recommendations in structured form is a significant leap towards seamlessly integrating CDS and EBM. A common type system (described in the next sub-section) that specifies the format for the structured form of the recommendations would be very helpful and makes it easy to not only extract, maintain and share, but also enable the development of various medical applications on the top of it.

The National Guideline Clearinghouse (NGC) is a widely used public resource for evidence-based clinical practice guidelines [14] that provides a large number of unstructured summaries containing information derived from guidelines by using a template of guideline attributes. The guidelines are categorized according to the topics of the disorder, treatment or intervention, and health service administration. The guidelines are further divided into various subtopics following a hierarchical structure that is derived from U.S. National Library of Medicine's (NLM) Medical Subject Headings (MeSH) [15].

Each guideline details its scope and methodology used to develop the guideline before presenting the relevant recommendations and evidence supporting the recommendations. Various other types of information such as the benefits/harms of implementing the recommendations, contraindications, and implementation are also included to comprehensively define the guideline. For our study, similar to UpToDate, we extracted the recommendations provided in these guidelines. An example of some of the recommendations as presented in the guideline is shown in Table 2. UpToDate and NGC are our two sources of clinical recommendations. In addition, we obtained the medical publications and articles that support these recommendations. We used these articles to evaluate the extensibility of our proposed type system for automated recommendation synthesis and thereby knowledge summarization and citation retrieval.

Table 1. Examples of Recommendations from UpToDate. The population group is underlined while the suggestion is highlighted in Italics and outcomes are in bold.

| Recommendations |
| --- |

We suggest an *ICD* for patients with Chagas cardiomyopathy (CCC) who survive an episode of sudden cardiac arrest or have sustained ventricular tachycardia, particularly if the patient was taking amiodarone and/or beta blocker therapy (Grade 2B).

We suggest *amiodarone plus beta blocker therapy* **to reduce shocks** in patients with Chagas cardiomyopathy (CCC) treated with implantable cardioverter defibrillator (ICD) (Grade 2C).

We suggest *amiodarone with or without beta blocker therapy* for patients with CCC, a Rassi score of ≥10, and no sustained VT on Holter (Grade 2C).

For patients with systolic HF and volume overload, we recommend *diuretics*.

For patients with HF with left ventricular systolic dysfunction (left ventricular ejection fraction [LVEF] ≤40 percent), we recommend *angiotensin converting enzyme (ACE) inhibitor therapy*.

Table 2. Examples of Recommendations from NGC. The population group is underlined while the suggestion is highlighted in Italics and outcomes are in bold.

| Recommendations |
|---|
| *Beta blocker treatment* is recommended in patients with HF and preserved LVEF who have: prior myocardial infarction; hypertension; and atrial fibrillation requiring control of ventricular rate |
| *Blood pressure monitoring* is recommended in patients with HF and preserved LVEF. |
| *Brain natriuretic peptide (BNP) and N-terminal pro-B-type natriuretic peptide (NTproBNP)* are useful in the diagnosis and prognosis of heart failure in patients with dyspnea of unknown etiology. |
| Counseling on the use of a *low-sodium diet* is recommended for all patients with HF, including those with preserved LVEF. |
| *ACE inhibitors* should be considered in all patients with HF and preserved LVEF who have symptomatic atherosclerotic cardiovascular disease or diabetes and one additional risk factor. |

## Type Systems for Information Extraction Algorithms

Information extraction deals with the task of extracting structured information from unstructured or semi structured data resources. This requires the use of natural language processing (NLP) or text-mining methods and algorithms to process free-form human language texts. In almost every information extraction algorithm, a template is predefined that consists of data elements required to be extracted from the respective documents. This template can be represented in many formats, with examples being a list format, a hierarchy format, or an xml-based type system format. Many efforts have been made in the past to give a definite structure to solve the information extraction problem. Apache UIMA [28] is one such architecture.

Unstructured Information Management Architecture, or UIMA, is a framework to process and analyze unstructured information such as text, speech, or video. Analysis engines are the building blocks of the architecture, which annotates the unstructured information with descriptive attributes. An analysis engine can operate on the document individually or can be arranged as a pipeline of many analysis engines. The descriptive attributes annotated by the annotators (instances of analysis engines) are called analysis results. A type system defines the template for the analysis results, defining the types, features or attributes for a type, relationship among types, etc. In other words, a type system states the various types of objects that may be discovered in the document by using an analysis engine. These types then have certain features or properties to characterize them. As an example, Age, Gender, and Ethnicity are some features or properties of type Person. In biomedical domain, Jonnalagadda et al identified 52 data elements that are currently being extracted from different studies to automate systematic reviews [29]. To represent these data elements, a list-based template is the most commonly used representation. However, as more succinctly elaborated in the Discussion section the list-based format fails to model the relationship between extracted data elements and therefore is not expressive in nature. Thus, a type system-based template becomes an important component of the information extraction-based architecture as it facilitates more compact, comprehensive representation and sharing of extracted knowledge and thereby can be used in various other computer applications.

The main contributions through our proposed approach are twofold. First, we propose an expressive, definite, flexible, compact, comprehensive, and computable type system to effectively represent any given clinical recommendation. Second, our approach provides an abstract type system template that is extendible to represent any medical article and can be used in a variety of medical applications.

## Methods

To compile the list of relevant data elements and consequently define CRTS, we first analyzed 50 recommendations (25 each) for two of the most common cardiovascular diseases - heart failure (HF) and atrial fibrillation (AFib). These recommendations were randomly selected from 49 UpToDate topics (27 for HF and 22 for AFib). We then performed a qualitative analysis of each recommendation and simultaneously tagged each with the data elements deemed appropriate. The analysis was done separately by the authors RG and KR and discrepancies were resolved by consulting SRJ.

Next, all the data elements are compiled into an exhaustive list. We ensured that all the important information from the recommendation can be represented using the compiled data elements. Finally, the data elements are used to build an Apache UIMA-based common type system. CRTS organizes all the data elements into a definite pattern and conveys all the information provided by the recommendation in a structured way.

In the type system, we also use negation tags and XML expression covering Conjunctive Normal Form (CNF) and Disjunctive Normal Form (DNF) [58] to accommodate the variations of language used by clinicians. For example, in the recommendation "For patients with systolic HF and volume overload, we recommend diuretics," the population is described by two medical conditions that are joined by an *AND*. This detail is very essential for comprehensively covering the information conveyed by the recommendation.

Finally, to validate CRTS, we used it to express a large number of randomly selected recommendations corresponding to various other disorders. The diseases/conditions we considered in our experiments are diabetes, stroke, tuberculosis, lung cancer, HIV/AIDS, and coronary heart disease. To ensure the generalizability of our system, we also obtained and analyzed an equal random sample of recommendations from NGC. The total number of recommendations for each disease/condition considered is presented in Table 3. We performed a qualitative analysis of each recommendation, expressing it in CRTS format.

Table 3. Diseases/Conditions and Number of Recommendations Considered for Validation

| Disease/Condition | Number of Recommendations | Number of Articles |
|---|---|---|
| Diabetes | 20 | 30 |
| Stroke | 15 | 20 |
| Tuberculosis | 20 | 35 |
| Lung cancer | 25 | 40 |
| HIV/AIDS | 25 | 40 |
| Coronary artery disease | 20 | 30 |
| **Total** | **125** | **195** |

For further analysis, we also investigated representing the abstracts of medical research articles and citations in CRTS format. We took a large random sample of the abstracts for the articles cited in UpToDate and NGC. The articles are confined to the diseases/conditions for which we obtained the recommendation. The number of articles used for our analysis is shown in Table 3.

## Results
## Proposed type system

A type system defines a data structure of manually or automatically extracted types or features for representing information in structured format. We present a condensed depiction of CRTS in Figure 2. Based on our analysis, each clinical recommendation can be seen as a constituent of three important clinical types - Population, Suggestion, and Outcome. As an example, for the recommendation "For patients with systolic HF who do not tolerate ACE inhibitors, we recommend an angiotensin II receptor blocker (ARB) as an alternative that provides a similar survival benefit," "patients with systolic HF who do not tolerate ACE inhibitors" is the population, "angiotensin II receptor blocker (ARB)" is the suggestion, and "survival benefit" is the outcome. We listed additional examples in Tables 1 and 2. Not only did our analysis show that all recommendations consist of population, suggestion and outcome, but it also has face validity. A recommendation is expected to have a set of suggestions for a particular population group to achieve a particular outcome.

```xml
<recommendation>
    <population>
        <demographics>
            <age>Numeric</age>
            <gender>Categorical</gender>
            <ethnicity>Categorical</ethnicity>
            <country>Categorical</country>
        </demographics>
        <disorder>
            <id>Numeric</id>
            <name>Text</name>
            <UMLSDictId>Numeric</UMLSDictId>
            <timeperiod>Categorical</timeperiod>
            <negation>Boolean</negation>
        </disorder>
        <intervention>
            <id>Numeric</id>
            <name>Text</Name>
            <type>Text</type>
            <dictionary>Text</dictionary>
            <dictId>Numeric</dictId>
            <timeperiod>Categorical</timeperiod>
            <grade>Categorical</grade>
        </intervention>
        <labResults>
            <id>Numeric</id>
            <key>Text</key>
            <value>Numeric</value>
            <operator>Categorical</operator>
            <temporal>Text</temporal>
        </labResults>
        <expr type=""> #type of the expression can be OR AND
            <inputConceptId>List</inputConceptId>
            <outputConceptId>Numeric</outputConceptId>
        </expr>
    </population>
    <suggestion>
        <intervention>
            <id>Numeric</id>
            <name>Text</Name>
            <type>Text</type>
            <dictionary>Text</dictionary>
            <dictId>Numeric</dictId>
            <timeperiod>Categorical</timeperiod>
            <grade>Categorical</grade>
        </intervention>
        <expr type=""> #type of the expression can be OR AND
            <inputConceptId>List</inputConceptId>
            <outputConceptId>Numeric</outputConceptId>
        </expr>
    </suggestion>
    <outcome>
        <generalOutcome>
            <id>Numeric</id>
            <outcomeText>Text</outcomeText>
        </generalOutcome>
        <labResults>
            <id>Numeric</id>
            <key>Text</key>
            <value>Numeric</value>
            <operator>Categorical</operator>
            <temporal>Text</temporal>
        </labResults>
    </outcome>
</recommendation>
```

**Figure 2.** XML based condensed description of CRTS.

These types are described by one or more subtype such as Demographics, Disorder, Intervention, Lab Results, Study Design, and General Output. Each of these subtypes have properties or features that form the characteristics of the object. We identified several essential data elements and features required to represent any given Recommendation. We summarize these key data elements that we have used to build CRTS in Table 4. More data elements, as needed, can be extracted and used as either features or clinical subtypes under the three major types of Population, Suggestion, and Outcome to represent a clinical recommendation in the type system.

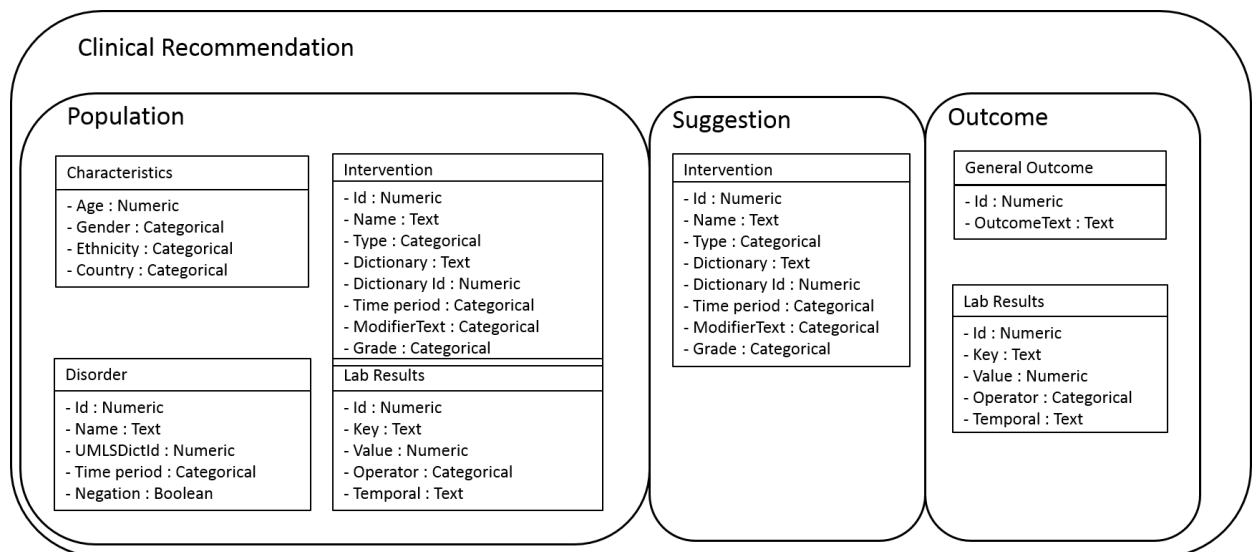

**Figure 3.** A visually representation of CRTS.

Figure 3 shows a visual representation of CRTS. In this figure, "Recommendation" is at the top of chart which is described by three major clinical concepts or types: Population, Suggestion, and Outcome. These concepts are then defined by other entities or clinical subtypes such as Demographics, Disorder, Interventions and Lab Results. Each of the subtype is characterized by specific features; for example, Demographics has the features of Age, Ethnicity, Gender, and Country. This can easily be extended to include more features depending on the domain used. The XML files for the UIMA type system are publicly available. <<Citation - system would be made publicly available upon acceptance of the publication as a reference here

\>>

Table 4. List of data elements we consider to design our type-system.

|  | Key Data Element | Subtype | Major Type(s) | Value Type |
|---|---|---|---|---|
| 1 | Age | Demographic | Population | Numeric |
| 2 | Gender | Demographic | Population | Categorical |
| 3 | Ethnicity | Demographic | Population | Categorical |
| 4 | Country | Demographic | Population | Boolean |
| 5 | Disorder name | Disorder | Population | Text |
| 6 | Disorder UMLS dict id | Disorder | Population | Numeric |
| 7 | Disorder negation | Disorder | Population | Boolean |
| 8 | Disorder time period | Disorder | Population | Categorical |
| 9 | Intervention Name | Intervention | Population/Suggestion | Text |
| 10 | Intervention Type | Intervention | Population/Suggestion | Categorical |
| 11 | Intervention time period | Intervention | Population/Suggestion | Categorical |
| 12 | Intervention dictionary | Intervention | Population/Suggestion | Text |
| 13 | Intervention modifier text | Intervention | Population/Suggestion | Categorical |
| 14 | Intervention Dict id | Intervention | Population/Suggestion | Numeric |
| 15 | Intervention Grade | Intervention | Population/Suggestion | Categorical |
| 16 | Lab result key | Lab Results | Population/Outcome | Text |
| 17 | Lab result value | Lab Results | Population/Outcome | Numeric or Range |
| 18 | Lab result operator | Lab Results | Population/Outcome | Categorical |
| 19 | Lab result temporal | Lab Results | Population/Outcome | Text |
| 20 | Outcome Text | General Outcome | Outcome | Text |

## Type system major types - Population, Suggestion, and Outcome
### *Population*

In clinical terms, population is defined as a group of people sharing the same set of characteristics and problem(s) [39]. This definition does not provide the details of various data elements that are used to define a population group. To reduce any ambiguity, in CRTS we concretely define the various data elements we use to represent a population group. These data elements are sub-divided into four subtypes: Demographics, Problem, Intervention, and Lab Results.

First, all patients in the population group must share the same demographics. The demographics used often in clinical recommendations include Age, Gender, Ethnicity, and Country. In CRTS, we include these data elements as features in the clinical subtype referred to as Demographics. Second, a population group is characterized by the medical condition(s) or the disorder(s). The recommendation may also require the patient group to either currently have a certain disorder or to have had it in the past. This detail needs to be correctly modeled in the type system in order to completely describe clinical evidence. In addition, the disorder text must be normalized to a concept in a medical taxonomy to eliminate cases of synonyms and language complexities. We use the UMLS Metathesaurus as an example in CRTS; however, any available medical taxonomy can be used. To summarize, the Disorder subtype is therefore characterized by the features: Id (unique id given to object block), Text (the disorder text), ConceptDictId (UMLS dict id), and Time-period (whether the disorder is current or occurred in the past). Third, a population group may also be characterized by any treatment including drugs, patients are being administered. Also, some recommendations may need the patients to have undergone a therapy in the past. We include this information in the Intervention clinical subtype. The information related to Intervention (treatment, therapy, drug or medicine) is depicted in conceptType feature. The text span must also be normalized to a specific concept. We use UMLS for the normalization, and the corresponding dictionary id is depicted in conceptDictId feature. Last, patients may also be characterized by certain laboratory results or measurements, such as hemoglobin, Rassi score, or left ventricular ejection fraction (LVEF) value. We account this information in subtype called Lab Results. The features we consider in this subtype include Keys, Values, Operator, and Temporal Information. Keys corresponds to the result name, Value to the actual result value that can be in a definite integer value format or as a range, Operator to the logical code, and Temporal to the time period of the corresponding lab result. For example, in "LVEF ≤ 40%," "LVEF" is the Key, "40%" is the Value, and "≤" is the Operator.

## Suggestion

Suggestion is the second main constituent of the Recommendation. Suggestion is the advice put forth by summarizing the evidence published in medical publications and articles for a particular patient group. Examples of Suggestion (in italics) can be found in Tables 1 and 2. In analyzing a large number of recommendations, we find that Suggestion constitutes Intervention with optional Comparison. We use the same Intervention subtype that we used for defining Population type including all its features and properties. In addition, in some recommendation multiple interventions may be suggested for certain conditions. We capture this information in form of specifically designed CNF (AND/Conjunction of ORs/Disjunctions) and DNF (OR/Disjunction of Ands/Conjunction) XML expression. We embed the comparison information in expression type attribute apart from *AND* and *OR* expressions.

## Outcome

Outcome, the expected result of following the Suggestion for the Population, is the third main constituent of the recommendation. Outcome can be quite general, such as "*to reduce shocks*" or "*to prevent further blood loss,*" or it can be specific in terms of lab results: "*to maintain digoxin* *levels between 0.5 and 0.8 ng/ml.*" In our current type system, we have two clinical subtypes to represent Outcome: General Output and Lab Values. General Output has the features Id and OutputText. Id is the concept id given to the block, while OutputText is the text-span denoting the output phrase. Lab Values subtype is the same subtype as described before for Population.

## Recommendation representation example

In Figure 4, we present an example to represent: "*We suggest amiodarone plus beta blocker therapy to reduce shocks in patients with Chagas cardiomyopathy (CCC) treated with implantable cardioverter defibrillator (ICDs) (Grade 2C)*" in the CRTS format. In this recommendation, "*patients with Chagas cardiomyopathy (CCC) treated with implantable cardioverter defibrillator (ICDs)*" is the Population group and "*amiodarone plus beta blocker therapy*" is the Suggestion, whereas "*to reduce shocks*" is the Outcome. For Population, patients must have the disorder "*Chagas cardiomyopathy*" as well as must have had the Intervention "*implantable cardioverter defibrillator.*" This highlights the advantages of our proposed type system. Merely extracting the disorder concept and intervention concept in a list format would not communicate the complete information conveyed by the recommendation. In a similar vein, our proposed type system is better than a plain PICO data element extraction because we break down each data element into atomic constituents that are easy to interpret and are in a more computable format. For Suggestion, clinicians have advised "*amiodarone*" and "*beta blocker therapy,*" In list format, this would be extracted as two different interventions and it would be difficult to comprehend the information conveyed. In CRTS, we have a specially tuned XML expression that properly aggregates the information and makes it easy to understand manually and compute automatically. Also, as depicted in the figure, all the text occurrences are normalized to a definite concept in a widely used medical ontology or dictionary.

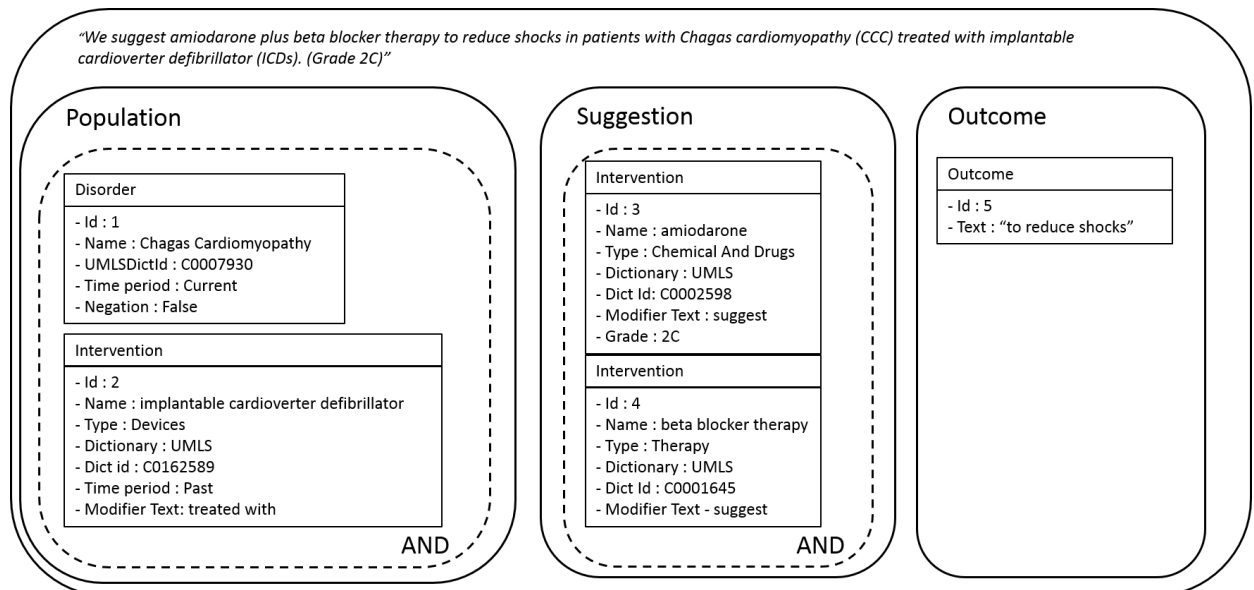

**Figure 4.** Representation example of a clinical recommendation.

In Multimedia appendix 1, we present some more examples of recommendations listed in Table 1 and Table 2 in the CRTS format.

## Medical article representation example

As mentioned earlier, Jonnalagadda et al identified 52 data elements that are commonly extracted in the systematic review process [29]. CRTS could be easily extended to include these data elements as either clinical types or features. The abstract we represent here is of the article "*Diuretics for heart failure*" [59] and is presented in textual form in Figure 5. In Figure 6, we produced a sample representation of a medical publication abstract in our proposed type system. As evident from Figure 5, the unstructured heterogeneous format is not only difficult to comprehend manually but also not in computable format. When represented in a type system that also automatically lends to a visual representation, it becomes easier and faster to comprehend and enables automated applications to be built on top of them. We discuss some examples of applications of our proposed type system in the Discussion section.

**Abstract**

**BACKGROUND:** Chronic heart failure is a major cause of morbidity and mortality world-wide. Diuretics are regarded as the first-line treatment for patients with congestive heart failure since they provide symptomatic relief. The effects of diuretics on disease progression and survival remain unclear.

**OBJECTIVES:** To assess the harms and benefits of diuretics for chronic heart failure

**SEARCH STRATEGY:** We searched the Cochrane Central Register of Controlled Trials (Issue 2 2004), MEDLINE 1966-2004, EMBASE 1980-2004 and HERDIN database. We hand searched pertinent journals and reference lists of papers were inspected. We also contacted manufacturers and researchers in the field.

**SELECTION CRITERIA:** Only double-blinded randomised controlled trials of diuretic therapy comparing one diuretic with placebo, or one diuretic with another active agent (e.g. ACE inhibitors, digoxin) in patients with chronic heart failure were eligible for inclusion.

**DATA COLLECTION AND ANALYSIS:** Two reviewers independently abstracted the data and assessed the eligibility and methodological quality of each trial. Extracted data were entered into the Review Manager 4.2 computer software, and analysed by determining the odds ratio for dichotomous data, and difference in means for continuous data, of the treated group compared with controls. The likelihood of heterogeneity of the study population was assessed by the Chi-square test. If there was no evidence of statistical heterogeneity and pooling of results was clinically appropriate, a combined estimate was obtained using the fixed-effects model.

**MAIN RESULTS:** We included 14 trials (525 participants), 7 were placebo-controlled, and 7 compared diuretics against other agents such as ACE inhibitors or digoxin. We analysed the data for mortality and for worsening heart failure. Mortality data were available in 3 of the placebo-controlled trials (202 participants). Mortality was lower for participants treated with diuretics than for placebo, odds ratio (OR) for death 0.24, 95% confidence interval (CI) 0.07 to 0.83; P = 0.02. Admission for worsening heart failure was reduced in those taking diuretics in two trials (169 participants), OR 0.07 (95% CI 0.01 to 0.52; P = 0.01). In four trials comparing diuretics to active control (91 participants), diuretics improved exercise capacity in participants with CHF, difference in means WMD 0.72 , 95% CI 0.40 to 1.04; P < 0.0001.

**AUTHORS' CONCLUSIONS:** The available data from several small trials show that in patients with chronic heart failure, conventional diuretics appear to reduce the risk of death and worsening heart failure compared to placebo. Compared to active control, diuretics appear to improve exercise capacity.

PMID: 16437464 [PubMed - indexed for MEDLINE]

Figure 5. Medical Publication Abstract.

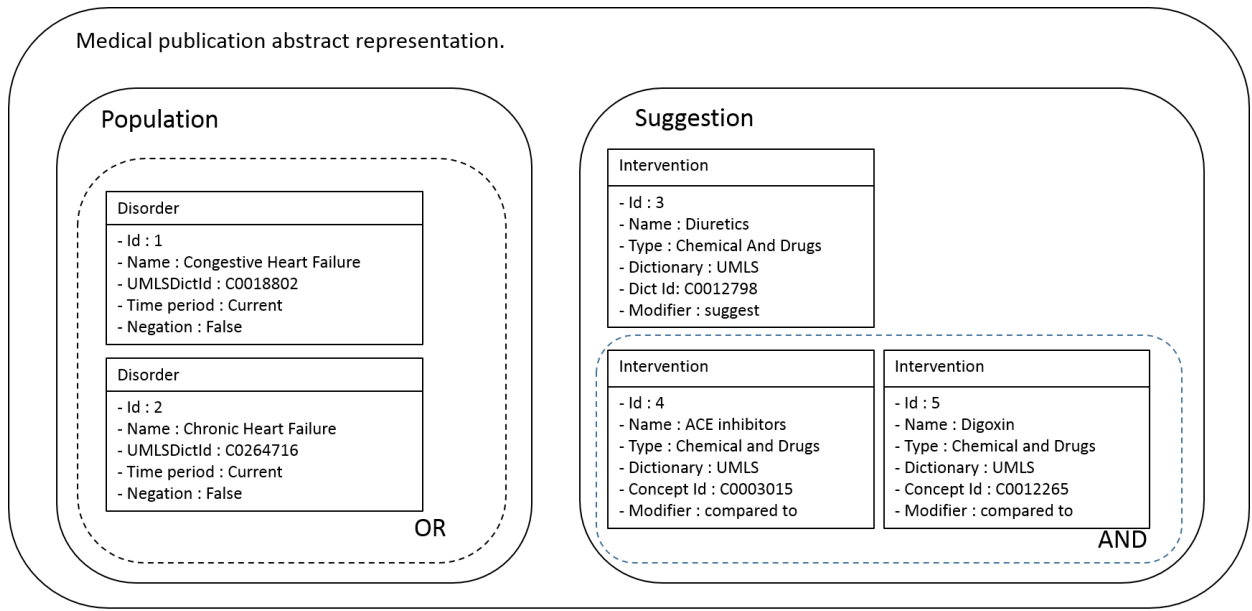

Figure 6. Representation of medical abstract in CRTS.

As an example, for a recommendation with no interventional comment or as we call it "Suggestion", consider the below abstract.

"*None of the available outcome-based studies was primarily designed to compare different blood pressure (BP) goals in patients with coronary artery disease (CAD). Consequently, there is uncertainty about the most appropriate BP treatment goal in these patients. Although US* *guidelines recommend a target less than 130/80 mm Hg, recent European guidelines state that such aggressive target is not consistently supported, therefore making the case for a less aggressive target (<140/90 mm Hg) in all hypertensive patients including those with CAD. A low systolic BP may be beneficial to limit myocardial workload, but an excessive lowering of diastolic BP might impair coronary perfusion, with potentially adverse effects (J-curve phenomenon). The optimal BP target for patients with CAD remains undefined. A reasonable target appears to be in the range of 130-140/80-90 mm Hg. Any further reduction may be safe, but not much productive from a prognostic standpoint.*"

There is no intervention suggested as to how to lower the blood pressure. In CRTS, the "population" type will represent that this abstract is about hypertensive patients including those with CAD, the "suggestion" type will be empty, and the "outcome" type will have "lab-result" subtype representing key "blood pressure" taking value of "140/90" as compared to value of "130/80" with operator has "<=" and units as "mm".

## Discussion

To review, we analyzed a large number of recommendations from UpToDate and NGC for various disorders, compiled commonly used data elements, and then used them to design a compact type system CRTS for effective representation. As demonstrated by our results, CRTS is able to represent clinical recommendations efficiently. We also showed how the basic type system can be extended to include various other data elements required to represent clinical knowledge in a medical publication abstract. We first show how CRTS is an enhancement over the previously proposed representation techniques in various aspects and then summarize the key features of CRTS. Finally we discuss some of the major applications where CRTS could be used.

## Data element list representation, Ontologies and existing Type systems

Some examples of list-based representations are PICO (Population, Intervention, Comparison and Outcome) [39], PECODR (Patient-Population-Problem, Exposure-Intervention, Comparison, Outcome, Duration, and Results) [40], PIBOSO (Population, Intervention, Background, Outcome, Study Design, and Other) [41], PICOT (Population, Intervention, Comparison, Outcome, and Time) [42], Cochrane Handbook [43], CONSORT (CONsolidated Standards of Reporting Trials) [44], and the STARD initiative (Standards for Reporting of Diagnostic Accuracy) [45].

Although list representations can be comprehensive, these have several practical issues. First, these lists do not define any relationships among the proposed data elements—for example, there may be multiple interventions suggested in a clinical recommendation, but this information is lost when converted to a list representation where the interventions would be present as two separate entities. Second, some of the data elements defined by these lists are ambiguous and do not generalize well. For example, a population group may also be characterized by an intervention that they may have taken in the past or are currently pursuing; however, these would be identified as interventions separately due to the ambiguity. In a similar vein, some lists (e.g., PICO, PIBOSO, PECODR, and PICOT) do not define specifically the data elements required to denote the types or subtypes. For example, the data elements required to denote the subtype Intervention include the type of intervention, the time of the intervention, and a dictionary id of extracted textual concept to mitigate multiple instances of same concept. This level of details is missing from the definition of PICO data elements. Similarly, Population is extracted as a phrase and thereby remain in highly unstructured form even after extraction. Other lists that define some of these specific features or attributes (e.g., Cochrane Handbook) do not define relationships between them and thereby fail to signify the complete meaning of the present information.

Example of these deficiencies can be seen in Figure 1, which represents a recommendation in these formats. As shown, in the PICO-based representation, Population is also characterized by an intervention, which would be extracted separately, remains in a highly unstructured form, and the relationship between the two interventions "amiodarone" and "beta blocker therapy," that they are both recommended simultaneously is not captured. Similar problems exist in other checklists such as the Cochrane Handbook, CONSORT- and STARD-based representations.

Our proposed type system differs from the ontological representation such as Gene Ontology and Human Phenotype Ontology [49, 50] used in UMLS that is present in some of the other information extraction systems such as MedLee (Medical Language Extraction and Encoding System) [51], ONYX [52] and MetaMap. The representation in these systems is a matching ontology concept along with the matched text-span and semantic type. Ontological representation is essential to normalize multiple instances of a concept but fails to define higher level concepts such as population, suggestion and outcome and capture other important features and properties such as age constraints, gender, ethnicity, country and lab values. Therefore, it cannot completely convey all the information contained in the recommendation. Further, they also fail to represent the conjunctions or disjunctions widely present between concepts in medical recommendations and guidelines.

Our work is similar in approach to the information extraction type systems used by biomedical NLP systems such as Apache cTAKES, JCoRe [53], HITEx [54], MedKAT [55], and U-Compare [56]. Most of these type systems are designed to work for information extraction systems tuned for data elements obtainable from electronic health records components such as pathology notes, diagnosis notes, progress notes, etc. For example, the cTAKES type system is based on UMLS concepts [57], providing a type system for six core SHARPn Clinical Element Models (Anatomical Sites, Disease and Disorders, Signs and Symptoms, Procedures, Medications, and Labs). However, it does not convey important information in recommendations and literature such as population, intervention, comparison and outcome. These systems do not provide the necessary types or data elements required to model clinical recommendations. For the same reason, it is difficult to represent medical publications and articles to effectively convey clinically actionable knowledge.

## Key Aspects of CRTS

**A. Expressive**: Existing representation techniques such as ontologies or type systems used by information extraction systems such as MetaMap do not define higher-level concepts such as population, suggestion and outcome, thereby failing to communicate the meaning behind clinical recommendation. Our XML-based type system, however, provides a highly compact structured form to the recommendations, clearly defining the important types, subtypes and features to symbolize clinical recommendations. As a result, CRTS is better suited to represent the information needs of clinicians and the data elements that need to be extracted by automated systems.

**B. Comprehensive and Definite:** As mentioned before in section 2.1, list based representations such as PICO, PICODR, PIBOSO, Cochrane Handbook, STARD, and CONSORT have ambiguous definitions of the concepts and lack the ability to properly capture all the aspects of clinical recommendation. CRTS effectively defines features or properties of each type or subtype and can be seen as an extension of these representations.

**C. Flexible:** As we have shown in Figure 1a and 1b, the existing representations are not flexible and do not completely define the relationship between extracted concepts. CRTS-based representation, however, enables the user to define a definite relationship between two or more extracted types or subtypes. They can be joined by mathematical operators such as *AND* and *OR* as well as more-complex relationships such as "*compared to*" and *"as opposed to."* In addition, it helps to handle negation effectively by using tag attributes. The type system framework is flexible for adding additional data elements and subtypes.

**D. Extendible:** CRTS not only is able to represent clinical recommendations effectively, but also is extendable to primary literature as shown in the Results section.

**E. Portable and Computable:** There is a high loss of information in list-based representation format. Higher-level applications such as CDS or knowledge summarization therefore perform poorly [60, 61]. As it is present in a computing friendly XML format, a type system-based representation is not only easy to share and port but also helps in building highly accurate medical applications.

## Applications
### *Knowledge Summarization*

Because of the exponential increase in the clinical studies and the number of publications that are available electronically, it is difficult for clinicians to keep abreast of the latest medical updates and automated approaches for knowledge summarization system are needed. However, due to the heterogeneity, complexity, and abundance of concepts in medical publications and articles, existing automated summarization techniques fail to give good results when directly applied to generate summaries. Existing approaches that summarize biomedical literature use vector similarity [63], concept matching [64], but very limited association extraction [65]. The accuracy or generalizability of the systems is highly limited. These issues can be solved by having an intermediate structured representation that can effectively symbolize the citation. As has already been shown in other fields, a structured representation plays a very important role in abstractive text summarization [66-68]. This structured representation, however, must be comprehensive enough to cover all the key information in the source document. CRTS is one such intermediate structured representation in this regard for summarizing medical literature and obtaining actionable knowledge in form of recommendations and guidelines. As we have shown, the data elements are exhaustive and effectively represent the information conveyed in the articles.

## *Citation retrieval*

Our proposed type system can also be extended and used to index citations. Citation retrieval has already been shown to work better using PICO framework than traditional information retrieval methods based on unstructured text [69-71]. CRTS based indexes would be more powerful as compared to PICO or other list formats based indexes since it provides a more compact structured form to each of the data elements. The indexed data could then be queried using XML Path [72] or other techniques.

## Limitations and Future Work

Although CRTS has several advantages as discussed earlier in this section, we have identified four limitations. First, the data elements we used to design the type system were compiled by analyzing recommendations for HF and AFib and then validated it for six other disease conditions. Nevertheless, the generalizability and comprehensiveness of the data elements covered in the current version need to be validated for several other disorders. We anticipate that this will be done prospectively and since our representation is based on type system format, it is easy to add new data elements in form of features or subtypes. Secondly, the primary goal of CRTS is to create a unified framework that can represent extracted data elements from various sources and be used as the backbone for various applications that rely on biomedical information extraction. Although, we demonstrated the usefulness of CRTS with such applications using a few examples, the implementation and usability has not yet been evaluated. In our future work, we will build upon and advocate the use of CRTS in CDS, knowledge summarization, citation retrieval, etc. Thirdly, CRTS is based on Apache UIMA and there is a large learning curve associated with using such a framework. We will build a Graphical User Interface (GUI) based on UIMA's own CAS Visual Debugger (CVD) for various configurations of our system. Finally, CRTS was built to represent clinically actionable knowledge in the form of Recommendations. We showed how this can be extended to represent medical publication abstracts. However, addition of new data elements may be required for more representing knowledge from other sources such as medical publication full-texts.

## Conclusion

We have presented CRTS, a common type system to represent clinical recommendations. We showed that our proposed type system is precise and comprehensive in representing a large sample of recommendations available for various diseases and conditions. We also list the data elements that are required to be extracted from the unstructured and heterogeneous text of recommendations to build the type system-based summary. We currently manually extract these required data elements; however, in future work, an automated system can be built for their extraction. We have also illustrated that the type system can also be used to represent recommendation information from scientific literature. The proposed type system-based structured summary can be used in a variety of medical applications such as CDSS, knowledge summarization, and citation retrieval.

## Acknowledgments

This work was made possible by funding from the National Library of Medicine grant:

R00LM011389.

## Abbreviations

AFib: Atrial Fibrillation

CDS: Clinical decision support

CRTS: Clinical recommendation type system

EBM: Evidence based medicine

HF: Heart Failure

ICD: Implantable cardioverter defibrillator

NLP: Natural language processing

# Multimedia Appendix 1

## Additional Examples of Recommendations represented in CRTS

"We suggest an ICD for patients with Chagas cardiomyopathy (CCC) who survive an episode of sudden cardiac arrest or have sustained ventricular tachycardia, particularly if the patient was taking amiodarone and/or beta blocker therapy.(Grade 2B)"

### Population

**Disorder**
- Id : 1
- Name : Chagas Cardiomyopathy
- UMLSDictId : C0007930
- Time period : Current
- Negation : False

**Disorder**
- Id : 2
- Name: Sudden cardiac arrest
- UMLSDictId : C1720824
- Time period: Past
- Negation : False

**Disorder**
- Id : 3
- Name: Sustained ventricular tachycardia
- UMLSDictId : C0750194
- Time period : Current
- Negation : False

OR

**Intervention**
- Id : 4
- Name : Amiodarone
- Dictionary : UMLS
- Dictionary id : C0002598
- Type : Chemicals & Drugs
- Time period : Past
- Modifier Text : "was taking"

**Intervention**
- Id : 5
- Name: Beta blocker
- Dictionary : UMLS
- Dictionary Id: C0001645
- Type: Chemicals & Drugs
- Time period: Past
- Modifier Text: "was taking"

AND/OR

AND

### Suggestion

**Intervention**
- Id : 6
- Name : ICD
- Type : Devices
- Dictionary : UMLS
- Dictionary id: C0002598
- Modifier : Suggest
- Grade : 2B

---

"For patients with HF with left ventricular systolic dysfunction (left ventricular ejection fraction [LVEF] ≤40 percent), we recommend angiotensin converting enzyme (ACE) inhibitor therapy. "

### Population

**Disorder**
- Id : 1
- Name : Heart failure
- UMLSDictId : C0018801
- Time period : Current
- Negation : False

**Disorder**
- Id : 2
- Name : Left ventricular systolic dysfunction
- UMLSDictId : C1277187
- Time period : Current
- Negation : False

**Lab value**
- Id : 3
- Key : Left ventricular ejection fraction
- Value range: 40
- Unit: percent
- Operator: <=

AND

### Suggestion

**Intervention**
- Id : 4
- Name : angiotensin converting enzyme (ACE) inhibitor therapy
- Type : Chemical And Drugs
- Dictionary : UMLS
- Dictionary Id : C0003015
- Modifier : Suggest

*"We suggest amiodarone with or without beta blocker therapy for patients with CCC, a Rassi score of ≥10, and no sustained VT on Holter."*

## Population

**Disorder**
- Id : 1
- Name : Chagas Cardiomyopathy
- UMLSDictId : C0007930
- Time period : Current
- Negation : False

**Lab value**
- Id : 3
- Name : Rassi score
- Value range: 10
- Operator: >=

**Disorder**
- Id : 2
- Name : Ventricular tachycardia on Holter
- UMLSDictId: C0042514
- Time period: Current
- Negation: True

AND

## Suggestion

**Intervention**
- Id : 4
- Name : beta blocker therapy
- Type : Therapy
- Dictionary : UMLS
- Dictionary Id : C0001645
- Modifier : suggest

**Intervention**
- Id : 5
- Name : Amiodarone
- Dictionary : UMLS
- Dictionary id : C0002598
- Type : Chemicals & Drugs
- Time period : Past
- Modifier : suggest

AND/OR

---

*"Beta blocker treatment is recommended in patients with HF and preserved LVEF who have: prior myocardial infarction; hypertension; and atrial fibrillation requiring control of ventricular rate ."*

## Population

**Disorder**
Id : 1
- Name : Heart failure
- UMLSDictId : C0018801
- Time period : Current
- Negation : False

**Disorder**
Id : 2
- Name : Left ventricular ejection fraction (LVEF)
- UMLSDictId : C0428772
- Time period : Current
- Negation : False

**Disorder**
- Name : Myocardial infarction
- UMLSDictId : C0027051
- Time period : Past
Id : 3

**Disorder**
Id : 4
- Name : Hypertension
- UMLSDictId : C0004238
- Time period : Current
- Negation : False

**Disorder**
Id : 5
- Name : Atrial fibrillation
- UMLSDictId : C0004238
- Time period : Current
- Negation : False

AND

## Suggestion

**Intervention**
Id : 6
- Name : Beta blockers
- Type : Therapy
- Dictionary : UMLS
- Dictionary Id : C0001645
- Modifier : recommended

*"Blood pressure monitoring is recommended in patients with HF and preserved LVEF ."*

## Population

**Disorder**

Id : 1
- Name : Heart failure
- UMLSDictId : C0018801
- Time period : Current
- Negation : False

**Disorder**

Id: 2
- Name: Left ventricular ejection fraction (LVEF)
- UMLSDictId: C0428772
- Time period: Current
- Negation : False

AND

## Suggestion

**Intervention**

- Id : 3
- Name : Blood pressure monitoring
- Type : Procedure
- Dictionary : UMLS
- Dictionary Id : C0242876
- Modifier : Suggest

---

*"In the absence of other specific indications for these drugs, angiotensin receptor blockers (ARBs) or angiotensin-converting enzyme (ACE) inhibitors may be considered in patients with HF and preserved LVEF."*

## Population

**Disorder**

Id : 1
- Name : Heart failure
- UMLSDictId : C0018801
- Time period : Current
- Negation : False

**Disorder**

Id: 2
- Name: Left ventricular ejection fraction (LVEF)
- UMLSDictId: C0428772
- Time period: Current
- Negation : False

AND

## Suggestion

**Intervention**

Id : 3
- Name : Angiotensin receptor blockers (ARBs)
- Type : Chemicals & drugs
- Dictionary : UMLS
- Dictionary Id : C0003015
- Modifier : may be considered

**Intervention**

Id : 4
- Name : Angiotensin-converting enzyme (ACE) inhibitors
- Type : Chemicals & drugs
- Dictionary : UMLS
- Dictionay Id : C0521942
- Modifier : may be considered

OR

*"Counseling on the use of a low-sodium diet is recommended for all patients with HF, including those with preserved LVEF."*

### Population

**Disorder**

Id : 1
- Name : Heart failure
- UMLSDictId : C0018801
- Time period : Current
- Negation : False

**Disorder**

Id: 2
- Name: Left ventricular ejection fraction (LVEF)
- UMLSDictId: C0428772
- Time period: Current
- Negation : False

AND

### Suggestion

**Intervention**

Id : 3
- Name : Low-sodium diet
- Type : Procedure
- Dictionary : UMLS
- Dictionary Id: C0012169
- Modifier : use of

---

*"ACE inhibitors should be considered in all patients with HF and preserved LVEF who have symptomatic atherosclerotic cardiovascular disease or diabetes and one additional risk factor. In patients who meet these criteria but are intolerant to ACE inhibitors, ARBs should be considered."*

### Population

**Disorder**

- Id : 1
- Name : Heart failure
- UMLSDictId : C0018801
- Time period : Current
- Negation : False

**Disorder**

- Id : 2
- Name : Left ventricular ejection fraction (LVEF)
- UMLSDictId : C0428772
- Time period : Current
- Negation : False

**Disorder**

- Id : 3
- Name: Atherosclerotic cardiovascular disease
- UMLSDictId : C0004153
- Time period : Current
- Negation : False

**Disorder**

- Id : 4
- Name : Diabetes
- UMLSDictId : C0011847
- Time period : Current
- Negation : False

OR

AND

### Suggestion

**Intervention**

- Id : 5
- Name : Angiotensin receptor blockers (ARBs)
- Type : Chemicals & drugs
- Dictionary : UMLS
- Dictionary Id: C0003015
- Modifier : should be considered if intolerant to ACE inhibitors

**Intervention**

Id : 6
- Name : Angiotensin-converting enzyme (ACE) inhibitors
- Type : Chemicals & drugs
- Dictionary : UMLS
- UMLSDictId : C0521942
- Modifier : should be considered

OR

"For patients with systolic HF and volume overload, we recommend diuretics."

Population

Disorder
- Id : 1
- Name : Systolic heart failure
- UMLSDictId : C1135191
- Time period : Current
- Negation : False

Disorder
- Id : 2
- Name: Volume overload
- UMLSDictId: C0546817
- Time period: Current
- Negation : False

AND

Suggestion

Intervention
- Id : 3
- Name : Diuretics
- Type : Chemical And Drugs
- Dictionary : UMLS
- Dictionary Id : C0012798
- Modifier : Suggest